\documentclass[a4paper]{article}

\usepackage{INTERSPEECH2022}

\title{indic-punct: An automatic punctuation restoration and inverse text normalization framework for Indic languages}
\name{Anirudh Gupta$^1$, Neeraj Chhimwal$^1$, Ankur Dhuriya$^1$, Rishabh Gaur$^1$, Priyanshi Shah$^1$, Harveen Singh Chadha$^1$, Vivek Raghavan$^2$}
\address{
  $^1$Thoughtworks\\
  $^2$Ekstep Foundation}
\email{anirudh.gupta@thoughtworks.com, vivek@ekstep.com}

\begin{document}

\maketitle
\begin{abstract}
Automatic Speech Recognition (ASR) generates text which is most of the times devoid of any punctuation. Absence of punctuation is text can affect readability. Also, down stream NLP tasks such as sentiment analysis, machine translation, greatly benefit by having punctuation and sentence boundary information. We present an approach for automatic punctuation of text using a pretrained IndicBERT model. Inverse text normalization is done by hand writing weighted finite state transducer (WFST) grammars. We have developed this tool for $11$ Indic languages namely Hindi, Tamil, Telugu, Kannada, Gujarati, Marathi, Odia, Bengali, Assamese, Malayalam and Punjabi. All code and data is publicly. available.\footnote{https://github.com/Open-Speech-EkStep/indic-punct}
\end{abstract}

\section{Introduction}

An ASR system usually produces an unpunctuated stream of text. Absence of punctuation can greatly affect readability. \cite{tundik2018user} showed that having no punctuation has a similar effect of having high word error rate. Restoring punctuations manually is a time consuming task. 

Prior approaches to automatic punctuation have used lexical and prosody features, or a combination of both. Some recent methods used acoustic features like pause duration, pitch and intensity and Hidden Markov Models (HMMs) to restore punctuation \cite{christensen2001punctuation}. These approaches even work better when combined with textual data. \cite{liu2006study, kolar2012development} proposed various ways to combine lexical features with prosody information and thereby improving punctuation restoration (PR) tasks. \cite{tilk2015lstm, tilk2016bidirectional} proposed unidirectional and bi-directional long short term memory networks which did not require hand made features. Recently the use of transformer based approaches with a combination of pre-trained word em-beddings have achieved state of art performance. \cite{yi2020adversarial} used pretrained BERT \cite{devlin2018bert} to restore punctuation. \cite{alam2020punctuation} studied various transformer architectures for PR and used an augmentation strategy which make the models more robust to ASR errors. Since we are training models for Indic languages we use IndicBERT \cite{kakwani2020indicnlpsuite} as the backbone and train PR models as a sequence labelling task.

Many times it is desired that the written form of put words is different from their spoken form. For example, somebody might speak an abbreviation but want the full form in the transcript or sometimes one can speak \textit{one thousand two hundred and for} but prefer 1,204 written on the transcript. Treatment of numbers by an ASR system has been a challenge. If numbers are considered as labels in the speech recognition task, then for some numbers in spoken form the model might be able to learn the correct written form. But since number combinations are infinitely many getting correct written forms directly from ASR is not a good approach. Inverse text normalization (ITN) converts spoken domain ASR output to written domain text to improve readability. Since in many applications there is very low tolerance for such kind of errors most of the systems in production today are rule based. Here we build upon the work from \cite{zhang2021nemo}. It uses a Python for ITN using WFST grammers. The write and compile grammers \textit{Pynini} \cite{gorman-2016-pynini} is used, which is a Python toolkit built on top of \textit{Openfst}.

We are making the following contributions for PR and ITN on Indic languages: 

\begin{itemize}
	\item We are open sourcing \textit{train}, \textit{valid} and \textit{test} sets for each of the $11$ languages. This can further act as a benchmark for PR task on Indic langauges\footnote{https://github.com/Open-Speech-EkStep/vakyansh-models}. 
	\item Language specific PR models are also open sourced. Our technique poses PR as a sequence labelling task \footnote{https://github.com/Open-Speech-EkStep/punctuation-ITN}. 
	\item WFST grammers for all languages are also open sourced. These can be used easily and extended in the same language or other languages. 
\end{itemize}

\section{Framework Parts}
Our framework contains two parts - punctuation restoration and inverse text normalization. Both of these components can be used separately or in conjunction with each other. 
\subsection{Inverse Text Normalization}
The ITN API consists of these parts: 

\begin{itemize}
	\item \texttt{classify} - creates a linear automaton from the input string and composes it with the final classification WFST, which transduces numbers and inserts semantic tags.
	\item \texttt{parse} - parses the tagged string into a list of key-value items representing the different semiotic tokens.
	\item \texttt{generate recordings} - is a generator function which takes the parsed tokens and generates string serializations with different reorderings of the key-value items.
		\item \texttt{verbalize} - takes the intermediate string representation and composes it with the final verbalization WFST, which removes the tags and returns the written form.

\end{itemize}

The most important class is \textit{CardinalFst}, since we are doing ITN mostly for handling numbers. Other grammars are built on top of it. For \textit{CardinalFst} we define a minimal set of number mappings for digits, teens and ties and use \textit{pynini.string} file to build a transducer that is the union of several string-to-string transductions from a TSV file. The rest of the graph is built from these building blocks by first creating a sub graph that consumes all three digit numbers. This is then composed with other graphs that consume quantities like thousand, million, billion, and so forth. We also handle currency.

\subsection{Punctuation Restoration}

\subsubsection{Dataset Preperation}
We use the IndicCorp\footnote{https://indicnlp.ai4bharat.org/corpora/} which is one of the largest publicly available corpora in Indian languages. We utilised $11$ languages in the dataset to train models. The following steps were performed to create training data. 

\begin{itemize}
	\item Filter out lines which contain some punctuation. We considered only three punctuations for all languages namely sentence end, comma and question mark.
	\item All the lines are normalized using \textit{IndicNLP}. 
	\item Sometimes there were words present in other languages apart from the language for which data was intended. We remove those words but leave those lines as it is. This was done so that we could simulate output from a speech recognition model which misses words as well. 
	\item Once clean text is obtained, lines of text are prepared for data loading. Each word is associated with a token. There are $4$ tokens in total. 
	\subitem [BLANK] - If there is no punctuation after that word. 
	\subitem [END] - If the sentence ends after that word. 
	\subitem [COMMA] - If there is a comma after that word. 
	\subitem [QM] - If there is a question mark after that word. 
\end{itemize}

\begin{table}[th]
	\caption{Dataset Description}
	\label{tab:data}
	\centering
	\begin{tabular}{llll}
		\toprule
		\textbf{Language} &  \textbf{train} & \textbf{valid} & \textbf{test} \\
		& (millions) & (thousands) & (thousands) \\
		\midrule
		Assamese	 & $1.08$    & $10$	    & $10$          \\
		Bengali          & $9$         & $10$  	  & $10$           \\
		Gujarati 		 & $14.9$    & $15$  		& $15$  		  \\
		Hindi			  & $3.8$     & $10$      & $10$			 \\
		Kannada 	  & $14.9$    & $15$ 		 & $15$ 		     \\
		Malayalam	 & $14.9$    & $10$ 		& $10$				\\
		Marathi 		& $14.9$	& $15$			& $15$				\\
		Odia 			  & $5.9$ 	   & $10$		   & $10$ 				\\
		Punjabi			& $9.9$		 & $10$			& $10$ 					\\
		Tamil 			 & $9.9$	  & $10$		  & $10$				\\
		Telugu 			& $14.9$    & $15$			& $15$				\\ 
		
		\bottomrule
	\end{tabular}
	
\end{table}

In all the languages, \textit{blank} is the most common label and accounts for almost $84 - 90$ \% of labels for all languages. Sentence \textit{end} form about $3-10$\% of the labels, \textit{commas} from $2 - 5$\% and \textit{question marks} are less than $1$\% for all languages. The distribution in validation and test sets was kept the same. 

\subsubsection{Model Architecture and Training}

We posed punctuation restoration as a token classification task and used IndicBERT \cite{kakwani2020indicnlpsuite} for that downstream task. IndicBERT is a multilingual ALBERT \cite{lan2019albert} model trained on $12$ major Indian languages. It also has less parameters than BERT \cite{devlin2018bert} or XLM-R \cite{conneau2019unsupervised} making it more suitable for the use of real time usage. 

\subsection{Results}
We calculate macro F1 score across labels for all languages i.e., unweighted mean of F1 scores on labels on the test set. Since the labels in dataset are highly imbalanced, unweighted mean is preferred otherwise the statistic was biased towards the label with highest count. The scores vary from $0.77 - 0.86$ for all languages. 

\begin{table}[th]
	\caption{F1 scores}
	\label{tab:results}
	\centering
	\begin{tabular}{ll}
		\toprule
		\textbf{Language} &  \textbf{F1 score}  \\
		\midrule
		Assamese	 & $0.81$           \\
		Bengali          & $0.81$           \\
		Gujarati 		 & $0.85$ 	  \\
		Hindi			  & $0.81$     \\
		Kannada 	  & $0.82$        \\
		Malayalam	 & $0.77$   	\\
		Marathi 		& $0.86$			\\
		Odia 			  & $0.84$ 	 		\\
		Punjabi			& $0.86$				\\
		Tamil 			 & $0.77$			\\
		Telugu 			& $0.81$  	\\ 
		
		\bottomrule
	\end{tabular}
	
\end{table}

\begin{figure}[th]
	\centering
	\includegraphics[width=\linewidth]{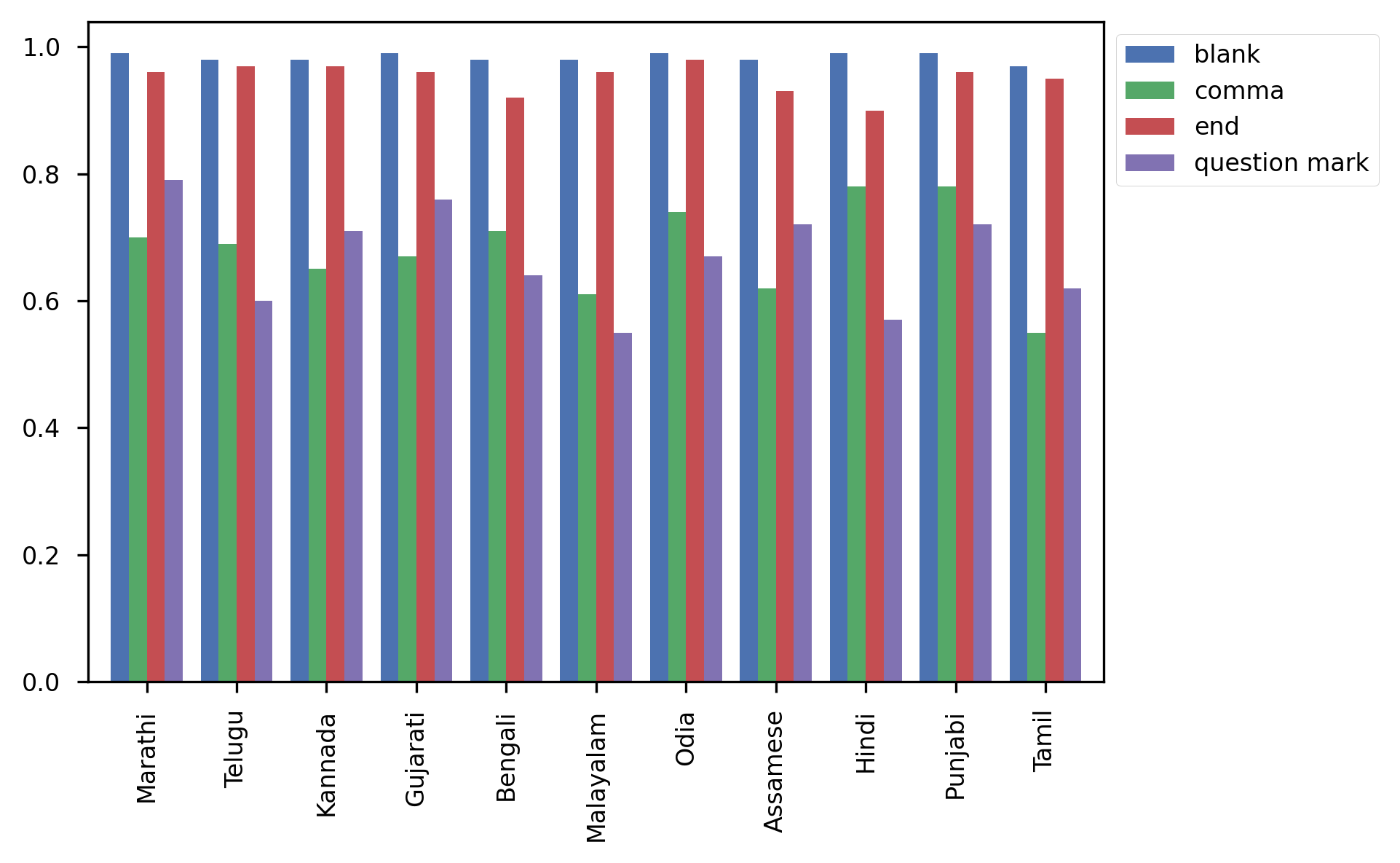}
	\caption{F1 scores across labels}
	\label{fig:results_labels}
\end{figure}

It is further observed that the F1 score for a particular label is directly proportional to the amount of times it appeared in the training data. It can be seen from Figure \ref{fig:results_labels} that \textit{blanks} are predicted most accurately, followed by \textit{end}, \textit{comma} and \textit{question mark} respectively. 

\section{Acknowledgements}
All authors gratefully acknowledge Ekstep Foundation for supporting this project financially and providing infrastructure. A special thanks to Dr. Vivek Raghavan for constant support, guidance and fruitful discussions. We also thank Ankit Katiyar, Heera Ballabh, Niresh Kumar R, Sreejith V, Soujyo Sen, Amulya Ahuja and Nikita Tiwari for helping out when needed and infrastructure support for data processing and model training.

\bibliographystyle{IEEEtran}

\bibliography{mybib}

\begin{thebibliography}{10}
\providecommand{\url}[1]{#1}
\csname url@samestyle\endcsname
\providecommand{\newblock}{\relax}
\providecommand{\bibinfo}[2]{#2}
\providecommand{\BIBentrySTDinterwordspacing}{\spaceskip=0pt\relax}
\providecommand{\BIBentryALTinterwordstretchfactor}{4}
\providecommand{\BIBentryALTinterwordspacing}{\spaceskip=\fontdimen2\font plus
\BIBentryALTinterwordstretchfactor\fontdimen3\font minus
  \fontdimen4\font\relax}
\providecommand{\BIBforeignlanguage}[2]{{%
\expandafter\ifx\csname l@#1\endcsname\relax
\typeout{** WARNING: IEEEtran.bst: No hyphenation pattern has been}%
\typeout{** loaded for the language `#1'. Using the pattern for}%
\typeout{** the default language instead.}%
\else
\language=\csname l@#1\endcsname
\fi
#2}}
\providecommand{\BIBdecl}{\relax}
\BIBdecl

\bibitem{tundik2018user}
M.~A. T{\"u}ndik, G.~Szasz{\'a}k, G.~Gosztolya, and A.~Beke, ``User-centric
  evaluation of automatic punctuation in asr closed captioning,'' 2018.

\bibitem{christensen2001punctuation}
H.~Christensen, Y.~Gotoh, and S.~Renals, ``Punctuation annotation using
  statistical prosody models.'' 2001.

\bibitem{liu2006study}
Y.~Liu, N.~V. Chawla, M.~P. Harper, E.~Shriberg, and A.~Stolcke, ``A study in
  machine learning from imbalanced data for sentence boundary detection in
  speech,'' \emph{Computer Speech \& Language}, vol.~20, no.~4, pp. 468--494,
  2006.

\bibitem{kolar2012development}
J.~Kol{\'a}r and L.~Lamel, ``Development and evaluation of automatic
  punctuation for french and english speech-to-text.'' in \emph{Interspeech},
  2012, pp. 1376--1379.

\bibitem{tilk2015lstm}
O.~Tilk and T.~Alum{\"a}e, ``Lstm for punctuation restoration in speech
  transcripts,'' in \emph{Sixteenth annual conference of the international
  speech communication association}, 2015.

\bibitem{tilk2016bidirectional}
------, ``Bidirectional recurrent neural network with attention mechanism for
  punctuation restoration.'' in \emph{Interspeech}, 2016, pp. 3047--3051.

\bibitem{yi2020adversarial}
J.~Yi, J.~Tao, Y.~Bai, Z.~Tian, and C.~Fan, ``Adversarial transfer learning for
  punctuation restoration,'' \emph{arXiv preprint arXiv:2004.00248}, 2020.

\bibitem{devlin2018bert}
J.~Devlin, M.-W. Chang, K.~Lee, and K.~Toutanova, ``Bert: Pre-training of deep
  bidirectional transformers for language understanding,'' \emph{arXiv preprint
  arXiv:1810.04805}, 2018.

\bibitem{alam2020punctuation}
T.~Alam, A.~Khan, and F.~Alam, ``Punctuation restoration using transformer
  models for high-and low-resource languages,'' in \emph{Proceedings of the
  Sixth Workshop on Noisy User-generated Text (W-NUT 2020)}, 2020, pp.
  132--142.

\bibitem{kakwani2020indicnlpsuite}
D.~Kakwani, A.~Kunchukuttan, S.~Golla, G.~N.C., A.~Bhattacharyya, M.~M. Khapra,
  and P.~Kumar, ``{IndicNLPSuite: Monolingual Corpora, Evaluation Benchmarks
  and Pre-trained Multilingual Language Models for Indian Languages},'' in
  \emph{Findings of EMNLP}, 2020.

\bibitem{zhang2021nemo}
Y.~Zhang, E.~Bakhturina, K.~Gorman, and B.~Ginsburg, ``Nemo inverse text
  normalization: From development to production,'' \emph{arXiv preprint
  arXiv:2104.05055}, 2021.

\bibitem{gorman-2016-pynini}
\BIBentryALTinterwordspacing
K.~Gorman, ``{P}ynini: A python library for weighted finite-state grammar
  compilation,'' in \emph{Proceedings of the {SIGFSM} Workshop on Statistical
  {NLP} and Weighted Automata}.\hskip 1em plus 0.5em minus 0.4em\relax Berlin,
  Germany: Association for Computational Linguistics, Aug. 2016, pp. 75--80.
  [Online]. Available: \url{https://aclanthology.org/W16-2409}
\BIBentrySTDinterwordspacing

\bibitem{lan2019albert}
Z.~Lan, M.~Chen, S.~Goodman, K.~Gimpel, P.~Sharma, and R.~Soricut, ``Albert: A
  lite bert for self-supervised learning of language representations,''
  \emph{arXiv preprint arXiv:1909.11942}, 2019.

\bibitem{conneau2019unsupervised}
A.~Conneau, K.~Khandelwal, N.~Goyal, V.~Chaudhary, G.~Wenzek, F.~Guzm{\'a}n,
  E.~Grave, M.~Ott, L.~Zettlemoyer, and V.~Stoyanov, ``Unsupervised
  cross-lingual representation learning at scale,'' \emph{arXiv preprint
  arXiv:1911.02116}, 2019.

\end{thebibliography}

% \begin{thebibliography}{9}
% \bibitem[1]{Davis80-COP}
%   S.\ B.\ Davis and P.\ Mermelstein,
%   ``Comparison of parametric representation for monosyllabic word recognition in continuously spoken sentences,''
%   \textit{IEEE Transactions on Acoustics, Speech and Signal Processing}, vol.~28, no.~4, pp.~357--366, 1980.
% \bibitem[2]{Rabiner89-ATO}
%   L.\ R.\ Rabiner,
%   ``A tutorial on hidden Markov models and selected applications in speech recognition,''
%   \textit{Proceedings of the IEEE}, vol.~77, no.~2, pp.~257-286, 1989.
% \bibitem[3]{Hastie09-TEO}
%   T.\ Hastie, R.\ Tibshirani, and J.\ Friedman,
%   \textit{The Elements of Statistical Learning -- Data Mining, Inference, and Prediction}.
%   New York: Springer, 2009.
% \bibitem[4]{YourName17-XXX}
%   F.\ Lastname1, F.\ Lastname2, and F.\ Lastname3,
%   ``Title of your INTERSPEECH 2022 publication,''
%   in \textit{Interspeech 2022 -- 23\textsuperscript{rd} Annual Conference of the International Speech Communication Association, September 18-22, Incheon, Korea, Proceedings, Proceedings}, 2022, pp.~100--104.
% \end{thebibliography}

\end{document}